\documentclass[preprint,12pt,authoryear]{elsarticle}

\usepackage{hyperref}       
\usepackage{url}            
\usepackage{booktabs}       
\usepackage{amsfonts}       
\usepackage{nicefrac}       
\usepackage{microtype}      
\usepackage{lipsum}
\usepackage{graphicx}

\graphicspath{ {./ }}
\usepackage{graphicx}
\usepackage{subcaption}
\usepackage{amsmath}
\usepackage{xcolor}
\usepackage{nomencl}
\makenomenclature

\journal{Mechanical Systems and Signal Processing}

\begin{document}

\maketitle
\begin{frontmatter}


\title{Learning the Helmholtz equation operator with DeepONet for non-parametric 2D geometries}





\author[aff1]{Rodolphe Barlogis}
\author[aff2]{Ferhat Tamssaouet}
\author[aff1]{Quentin Falcoz}
\author[aff1]{Stéphane Grieu}

\affiliation[aff1]{organization={PROMES-CNRS, Université Perpignan Via Domitia (UPVD)},
            city={Perpignan},
            country={France}}

\affiliation[aff2]{organization={LAAS-CNRS, Université de Toulouse (UT)},
            city={Toulouse},
            country={France}}

\begin{abstract}

This paper deals with solving the 2D Helmholtz equation on non-parametric domains, leveraging a physics-informed neural operator network, the DeepONet framework. We consider a 2D square domain with an inclusion of arbitrary boundary geometry at its center. It acts as a scatterer for an incoming harmonic wave. The aim is to learn the operator linking the geometry of the scatterer to the resulting scattered field. A signed distance function to the boundary of the inner inclusion evaluated in several points on the domain is used to encode its geometry. It serves as input for the branch part of the DeepONet architecture and local information as the input for the trunk part. This approach enables the encoding of arbitrary geometries, whether they are parameterized or not. The evaluation of the model on unseen geometries was compared to its finite element method (FEM) equivalent to test its  generalization capabilities. The trained network weights implicitly embed the local physics and their interaction with the domain geometry. If the training space sufficiently covers the target evaluation space, the model can generalize accordingly. Furthermore, it can be refined to extend to another region of interest without retraining from scratch. This framework also avoids the need to remesh the domain for each geometry. The proposed approach delivers a computationally lighter surrogate model than FEM alternatives and avoids relying on FEM generated training data.

\end{abstract}

\begin{graphicalabstract}
\centering
\includegraphics[width=\textwidth]{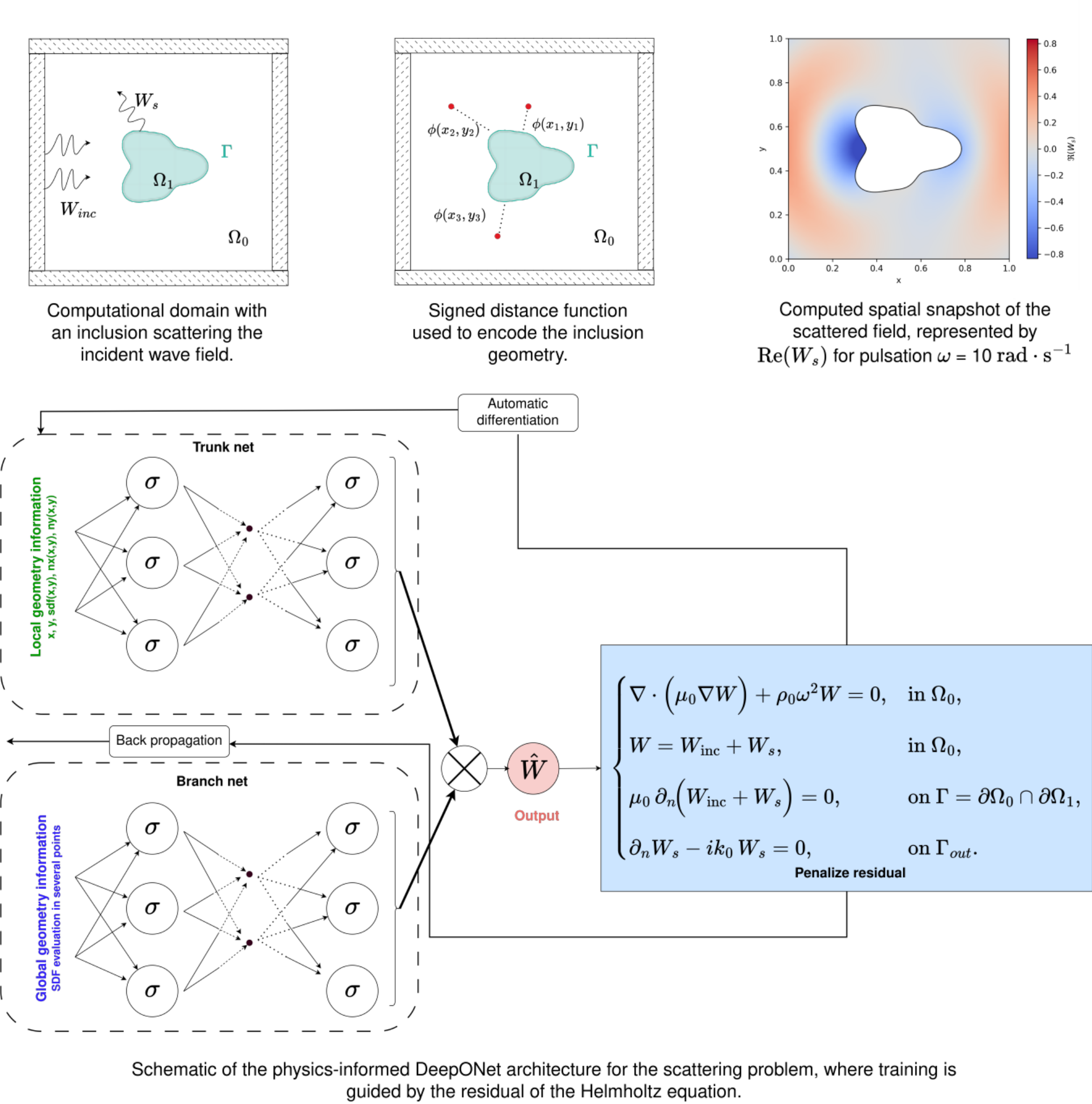}
\end{graphicalabstract}

\begin{highlights}
\item Representation of non-parametric geometries through a signed distance function (SDF) encoding,
\item Learning without FEM-generated data, relying solely on a physics-based loss function (i.e., effectively unsupervised),
\item Competitive evaluation against FEM solutions, with advantages such as fast inference and no need for remeshing,
\item Demonstration of generalization capabilities within a certain range of geometries.
\end{highlights}


\end{frontmatter}

\nomenclature{$\lambda_m$}{The first Lamé parameter of the domain m material}
\nomenclature{$\mu_m$}{The second Lamé parameter ; shear modulus of the domain m material}
\nomenclature{$k$}{Wave number}
\nomenclature{$\Omega_0$}{Whole domain except the inclusion}
\nomenclature{$\Omega_1$}{Inner inclusion domain}
\nomenclature{$\Gamma$}{Inner inclusion boundary}
\nomenclature{$\Gamma_{out}$}{Outer border boundary}
\nomenclature{$W_{inc}$}{Incident field}
\nomenclature{$W_{s}$}{Scattered field}
\nomenclature{$|W_m(x,y)|$}{Local oscillation amplitude}
\nomenclature{$\phi(x,y)$}{Signed radial distance function to the inner inclusion boundary evaluated in $(x,y)$}
\nomenclature{PINN}{Physics-informed neural network}
\nomenclature{AD}{Automatic differentiation}
\nomenclature{PDE}{Partial Differential Equation}
\nomenclature{BIE}{Boundary integral equation}
\nomenclature{GDF}{Geodesic distance function}
\nomenclature{FEM}{Finite element method}
\nomenclature{SDF}{Signed distance function}

\clearpage
\printnomenclature

\section{Introduction}

For problems governed by partial differential equations (PDE), difficulty is that data are rarely available in large quantities at low cost. In most cases, generating training data requires either expensive experiments or repeated high-fidelity numerical simulations for many geometries, source terms, and boundary conditions. This makes the construction of large datasets computationally demanding, and motivates the search for lightweight surrogate models that can alleviate both data and computational burdens. This need is especially acute when repeated evaluations across multiple geometries are required, since traditional FEM solvers may then become a severe bottleneck.  

In this case, operator learning, especially when combined with automatic differentiation, provides a natural framework for addressing PDE-driven problems. Rather than approximating pointwise input–output mappings, neural operators aim to learn the underlying solution operator itself, i.e., a mapping from functional inputs—such as distributed forces, boundary conditions, or geometric descriptors—to full solution fields. This paradigm is particularly well suited for problems where the input space is inherently infinite-dimensional. In the present work, we adopt this perspective to learn the operator that maps the geometry of a domain to the corresponding solution fields of the Helmholtz equation.

\subsection{Related works}

The present work follows a growing body of research investigating the use of neural operators to recover PDE solutions as a function of geometry. A central challenge in this line of research lies in the representation of geometry and its integration into operator learning frameworks.

~\cite{nairPhysicsGeometryInformed2025} propose a physics- and geometry-informed Deep Operator Network (DeepONet) for the efficient simulation of acoustic scattering (obstacles) by arbitrarily shaped rigid obstacles. Their approach combines a physics-informed training strategy, where the Helmholtz equation and boundary conditions are enforced directly in the loss function, with a non-uniform rational B-spline (NURBS) parameterization of the geometry. In this framework, geometric information is encoded in the branch network through a compact set of NURBS control points, while spatial coordinates are processed by the trunk network. This low-dimensional parametrization alleviates the curse of dimensionality typically associated with geometric inputs and contributes to training stability. The architecture relies on ResNet blocks for both the branch and trunk networks.

Applied to two-dimensional rigid body acoustic scattering, it learns a solution operator that maps scatterer geometry to the corresponding pressure field. Numerical results demonstrate high accuracy for both circular and arbitrary shapes that were not seen during training (within the NURBS family and over the range of parameters chosen for the control points), with relative $L_2$ errors typically below 15\% and $R^2$ scores exceeding 0.97 when compared to finite element (FE) solutions. However, performance degrades for geometries exhibiting sharp features, highlighting an intrinsic limitation of spline-based representations. Indeed, while NURBS provide a flexible and widely adopted framework for geometric modeling, they inherently produce smooth curves and surfaces, making them less suitable for capturing discontinuities, corners, or highly non-smooth features. Additionally, the computation of boundary normals—required to enforce interface conditions—is not explicitly detailed in this study, which raises practical questions regarding implementation.

This line of work has been pursued in \cite{nairMultipleScatteringSimulation2025} with configurations involving multiple scatterers. The authors leverage the linearity of the Helmholtz equation to superimpose the scattered fields generated by individual obstacles. Specifically, the domain is decomposed according to the number of scatterers, and each subdomain is associated with a network whose output represents the contribution of the corresponding scatterer to the total scattered field. The total scattered field, obtained as the sum of these partial fields, is then inserted into the residual of the Helmholtz equation to evaluate the loss function. The resulting gradients are subsequently backpropagated so as to train all subnetworks simultaneously. However, the proposed methodology does not seem to allow for generalization. It is able to learn a specific forward problem, but the model must be retrained whenever a different domain or new scatterers are considered. Moreover, this approach relies on linear superposition and may not directly extend to nonlinear PDEs.

A different perspective is proposed by ~\cite{fangLearningOnlyBoundaries2024}, who introduce a physics-informed neural operator framework based on boundary integral formulations. Instead of learning over the full domain, the method restricts the learning process to the boundary, where unknown potentials are inferred. By reformulating PDEs as boundary integral equations (BIEs), the learning process focuses on unknown boundary potentials while implicitly satisfying the governing equations inside and outside of the domain. Using boundary integral representations, the proposed method reduces the dimensionality of the learning problem from  $\mathcal{O}(N^d)$ to $\mathcal{O}(N^{d-1})$, where $d$ is the spatial dimension. This results in significantly lower training cost compared to conventional physics-informed neural networks (PINNs), which require dense interior collocation points. The approach is implemented using a neural operator architecture based on Nonlinear Manifold Decoders (NOMAD), which learns a mapping from geometric parameters and boundary observations to boundary potentials. An advantage of the method is its ability to handle complex and parameterized geometries as well as unbounded domains. Once the boundary potentials are inferred, solutions can be evaluated anywhere in the domain, including the far field, via the integral representations. The authors demonstrate the effectiveness of the approach on several benchmark problems, including the 2D Laplace equation, the 2D bi-harmonic equation, and the 3D Helmholtz scattering problem, achieving low relative errors (below 5\%) when compared to boundary element method (BEM) reference solutions. To sum it up, the authors propose a boundary-integral neural operator framework that shifts learning from the full domain to the boundary, reducing computational cost while preserving accuracy on complex geometries and unbounded problems. Overall, this work establishes a connection between classical boundary integral methods and operator learning.

Several recent contributions focuse on enhancing the geometric awareness of neural operators. ~\cite{kimPRTDeepONetGeometryawareNeural2026} introduce PRT-DeepONet, a geometry-aware deep operator network designed to efficiently predict pore-scale concentration fields in heterogeneous porous media governed by advection–diffusion–reaction equations. Instead of relying solely on a convolutional encoder to extract geometric features from binary pore images, the authors add to the trunk network a geodesic distance function (GDF). Trained on lattice Boltzmann simulations covering a wide range of transport and reaction regimes, PRT-DeepONet achieves significantly improved geometric fidelity compared to \textit{vanilla} DeepONet and Porous-DeepONet baselines. The GDF follows the same idea as the signed distance function (SDF) introduced in the present paper (Section \ref{para:sdf}). This method relies on supervised learning from simulation data, thus inheriting the cost of dataset generation.

Similarly, ~\cite{peyvanFusionDeepONetDataefficientNeural2026} intend to propose a fast geometry dependent surrogate model for flow field problem in order to ease the development of hypersonic and supersonic vehicles. Geometry is encoded in parametric descriptor, which serves as input of the branch network. The network is trained on high-fidelity discontinuous Galerkin spectral-element (DGSEM) simulations. In the same vein, ~\cite{heGeomDeepONetPointcloudbasedDeep2024} propose full-field solutions on variable three-dimensional parameterized geometries. In addition to standard spatial coordinates, the authors augment each query point with a signed distance function (SDF) evaluated with respect to the external surface of the geometry. The branch input consists of a parameterization of the geometric features. In this work, the objective is to predict the three-dimensional stress field. However, the model is not trained solely by minimizing the residual of the governing stress-equilibrium equations; it also relies on reference data generated by FEM simulations. The approach should therefore be regarded as a supervised, data-driven learning framework rather than a purely physics-informed one. Moreover, unlike our proposed approach, the branch input used to encode the geometry is directly based  on its defining parameters. As a result, the range of geometries that can be covered is inherently limited. 

Beyond parametric representations, ~\cite{leeHybridIterativeSolvers2025} introduce Geo-DeepONet: a class of geometry-aware hybrid iterative solvers that combine classical numerical methods with neural operator preconditioners to solve parametric PDEs on arbitrary unstructured domains. The main contribution is the introduction of Geo-DeepONet, a geometry-aware Deep Operator Network designed to generalize across unstructured meshes without retraining. Unlike standard DeepONets, which encode spatial information only through coordinates, Geo-DeepONet explicitly incorporates mesh connectivity information extracted from finite-element discretizations. This is achieved through a graph-based trunk network. It processes nodal coordinates together with the adjacency matrix of the mesh, thereby embedding geometric and topological information directly into the operator representation.

Finally, ~\cite{parkPointDeeponetPredictingNonlinear2026} propose Point-DeepONet, a hybrid operator-learning architecture that integrates PointNet ~\cite{qi2017pointnet}, a neural network designed to process unordered point clouds via pointwise feature extraction and symmetric aggregation, into the DeepONet framework, thereby enabling high-fidelity surrogate modeling for nonlinear structural analysis on non-parametric three-dimensional geometries.
The geometric representation proposed in this work is based on a similar idea. The difference is that, in our approach, the geometry is encoded by evaluating a signed distance function (SDF) at points located away from the geometric boundary itself. Points evaluated do not need to cover the whole space, it is way lighter for the branch part. Moreover, in their framework, FEM simulation data are again required to guide the learning process. Without these data, the model would not be able to learn an operator, since the network would have almost no effective training signal to distinguish the responses associated with different geometries.

\subsection{On the advantages of PINNs for modeling}

The main appeal of physics-informed surrogate models, as opposed to full finite element simulations, lies in their potential for fast inference. This benefit, however, should not be confused with training efficiency. PINN training is computationally demanding because the governing equation residual and its derivatives must be evaluated at each collocation points, by means of automatic differentiation, and convergence may require a large number of iteration steps. In a systematic comparison with FEM, ~\cite{grossmannCanPhysicsinformedNeural2024}. concluded that PINNs did not outperform the finite element method in terms of solution time and accuracy, although they could be faster during the evaluation stage once training was completed.

This can be explained by the fact that the inference stage of a feedforward neural network has computational complexity
$O\left(\sum_{l=1}^{L} n_{l-1}n_l\right)$,
where $L$ is the number of layers and $n_l$ is the number of neurons in layer $l$. In other words, the dominant cost arises from the matrix-vector products performed between successive layers. These operations are highly parallelizable and therefore particularly well adapted to GPU execution.

In comparison, FEM can exhibits a similar complexity,  it depends on several factors, including the strategy used to assemble the algebraic system, the sparsity pattern of the matrix, and the linear solver considered. Furthermore, while FEM computations can also benefit from parallelization, they are generally less directly amenable to efficient large-scale parallel execution than neural-network inference.

Consequently, the initial training cost is worthwhile when it can be amortized across many evaluations of the learned model, as is the case in repeated-query scenarios such as design exploration, optimization, or inverse problem.

Another advantage of such models is that the network weights implicitly encode the imposed physical constraints. Moreover, a model that has already been trained can in some cases be fine-tuned rather than retrained from scratch, provided that the new problem remains sufficiently close to the original one in terms of geometry or physical conditions. Recent work, such as proposed by \cite{takaoFinetuningPhysicsinformedNeural2026} on fine-tuning PINNs reports improved convergence when the pretrained model is sufficiently similar to the target configuration 






\subsection{Contributions of this work}

Overall, most existing works rely, at least partially, on supervised learning using high-fidelity simulation data to learn an operator. In contrast, the present work aims to learn the Helmholtz solution operator solely from the governing equations. A comparable approach is explored by ~\cite{nairPhysicsGeometryInformed2025}, discussed in the previous subsection. The main difference lies in the geometry encoding, and consequently in the range of geometries that can be represented. The main contributions of this paper are:

\begin{itemize}
    \item Representation of non-parametric geometries through a signed distance function (SDF) encoding,
    \item Learning without FEM-generated data, relying solely on a physics-based loss function (i.e., effectively unsupervised),
    \item Competitive evaluation against FEM solutions, with advantages such as fast inference and no need for remeshing,
    \item Demonstration of generalization capabilities within a certain range of geometries.
\end{itemize}

\subsection{Paper outline}

Now that related work has been reviewed, the proposed methodology is presented in Section~\ref{sec:methodo}. This is followed by numerical experiments and validation in Section~\ref{sec:result}, before devoting a final section to conclusions and future perspectives (Section~\ref{sec:persp}).

\section{Methodology}\label{sec:methodo}

The aim is to learn the field scattered by a scatterer as a function of the domain geometry. To this end, the governing wave equation, the domain, and the physical parameters governing the problem are defined, as they provide the information required for training the model. The objective is to obtain a lightweight surrogate model with fast inference, whose predictions depend explicitly on the domain geometry. For this purpose, a PINN-DeepONet approach is proposed.

\subsection{The scattering problem}

This paper considers wave propagation in an elastic material. For isotropic solids, the governing motion model is given by the vector Navier equation :

\begin{equation}
\rho\,\frac{\partial^2 \mathbf{u_m}}{\partial t^2}
= \mu\,\Delta \mathbf{u_m} + (\lambda+\mu)\,\nabla(\nabla\cdot \mathbf{u_m}),    
\end{equation}
where $\lambda$ and $\mu$ are the Lamé parameters characterizing the isotropic material: $\lambda$ governs compressibility, while $\mu$ is the shear modulus that controls deviatoric deformation. Here, $\mathbf{u_m}$ denotes the displacement field in subdomain $\Omega_m$. This equation can be derived from a more general wave equation.

To simplify the problem, we restrict our attention to horizontally polarized shear (SH) waves, in which the displacement takes the form $\mathbf{u}_m(x,y,t) = (0,0,w_m(x,y,t))$. This choice decouples out-of-plane motion from in-plane longitudinal (P) and shear (SV) waves, avoiding mode conversion at interfaces. Under the time-harmonic assumption $w_m(x,y,t) = \Re({W_m(x,y)e^{-i\omega t}})$, and partitioning the domain into two subregions $\Omega_0$ (the surrounding medium) and $\Omega_1$ (the inclusion), the problem reduces to the Helmholtz equation in each region:

\begin{equation}
\nabla \cdot \big(\mu_m \nabla W_m\big) + \rho_m \omega^{2} W_m = 0,
\qquad \text{in } \Omega_m,\; m\in\{0,1\}.    
\end{equation}

Linearity of the equations is exploited to decompose the total field into an incident part $W_{inc}$ (propagating from infinity) and a scattered part $W_s$. In particular, the incident field is taken as a plane wave 
\begin{equation}
    W_{\mathrm{inc}} = |W_{inc}|\, e^{\,i k_{0}(\mathbf{d}\cdot\mathbf{x})},
\end{equation}
where $|W_{inc}|$ is the incident wave amplitude, $\mathbf{d}$ the propagation direction vector, and $k_0$ the wavenumber in $\Omega_0$. As $\Omega_1$ is considered as a hole (void) so $\mu_1 = 0$, $\lambda_1 = 0$ and $\rho_1 = 0$, implying that the incident wave is entirely reflected (there is no transmitted part nor absorbed) at the boundary: $\Gamma=\partial\Omega_1$. Therefore, to satisfy the continuity conditions, the following conditions are imposed at the interface $\Gamma = \partial \Omega_0 \cap \partial \Omega_1$:
\begin{equation}
\begin{cases}
W_0 = W_{\mathrm{inc}} + W_s,
& \text{in } \Omega_0, \\[8pt]
\mu_0\,\partial_n\!\big(W_{\mathrm{inc}} + W_s\big)=0,
& \text{in } \Gamma 
\end{cases}    
\end{equation}
where $\partial_n$ denotes the normal derivative on $\Gamma$. The latter condition enforces zero flux through the rigid boundary. On the outer boundary of the domain (the perimeter of the unit square), an absorbing (non-reflecting) boundary condition is imposed on the scattered field~$W_s$.

\begin{figure}[htbp]
    \centering
    \includegraphics[width=0.35\textwidth]{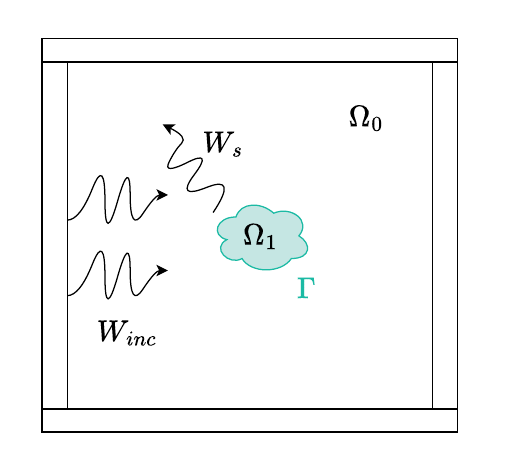}
    \caption{2D Domain with inclusion and absorbing outer borders. Highlight the incident wave field $W_{inc}$ and the scattered field $W_{s}$}
    \label{Fig:domain_representation_methodo}
\end{figure}

\ref{Fig:domain_representation_methodo} illustrates the schematic representation of the two-dimensional domain on which the proposed methodology is formulated.

\subsection{A PINN DeepOnet Framework}

\begin{figure} 
    \centering
    \includegraphics[width=0.95\textwidth]{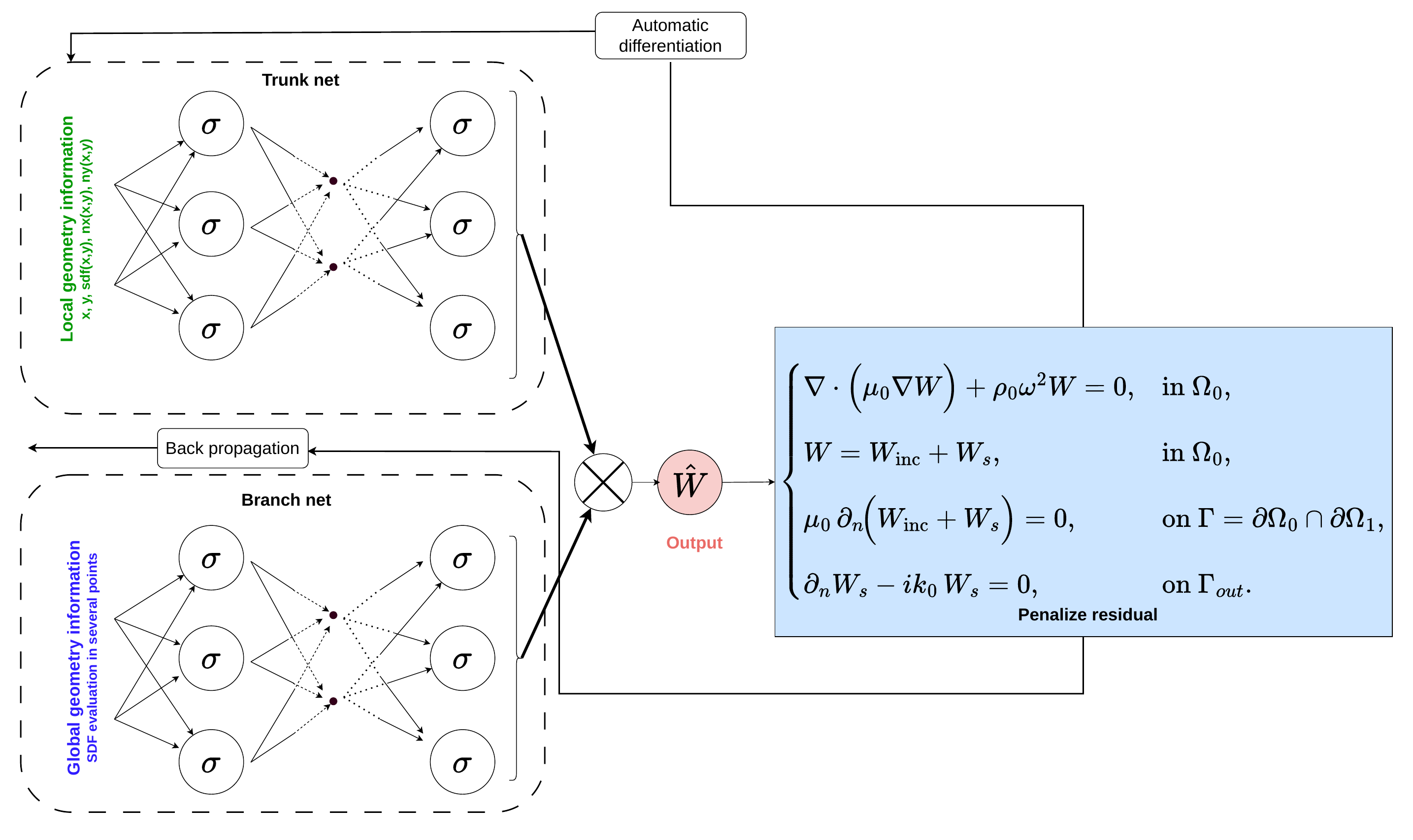}
    \caption{Schematic representation of the physics-informed DeepONet solving an Helmholtz problem. Highlight the decomposition of total wave field into the incident field $W_{inc}$ and the scattered field $W_{s}$}
    \label{Fig:deeponet_schematic}
\end{figure}

Physics-Informed Neural Networks (PINNs) is a technical concept revitalized by the seminal paper of ~\cite{raissiPhysicsinformedNeuralNetworks2019a} that aims to integrate physical laws directly into the deep learning architecture. It proposes to leverage AD \cite{baydinAutomaticDifferentiationMachine2018a} in order to compute the partial derivatives of neural network output with respect to its input coordinates (such as space and time). By backpropagating derivatives through the computational graph, PINNs can evaluate the residual of PDE at any given point. This residual is then incorporated into the loss function as a regularization term, penalizing solutions that violate the underlying physics. This same foundational principle, using AD to enforce physical constraints, is also a key component in the training of Deep Operator Networks (DeepONets) and other neural operator frameworks. DeepONets \cite{luDeepONetLearningNonlinear2021} are specialized neural architectures designed to learn mapping between infinite-dimensional function spaces, effectively approximating both linear and nonlinear operators. The architecture is composed of two distinct sub-networks: a branch net and a trunk net, as illustrated in Figure \ref{Fig:deeponet_schematic}.

In the context of this study, which aims to map geometries to the resulting solution fields of the Helmholtz equation, the network's inputs are partitioned by scale:
\begin{itemize}
    \item The branch net processes "global" information, encoding the overall geometric configuration or boundary conditions.
    \item The trunk net processes "local" information, typically the continuous coordinates $(x, y, z)$ where the solution is evaluated.
\end{itemize}
The final output is generated by computing a dot product between the latent representations of both networks, allowing the model to generalize across varying geometries rather than being limited to a single domain.

\paragraph{Evaluate a signed distance function $\phi$ as the input of the branch net}\label{para:sdf}

\begin{figure}[htbp]
    \centering
    \includegraphics[width=0.35\textwidth]{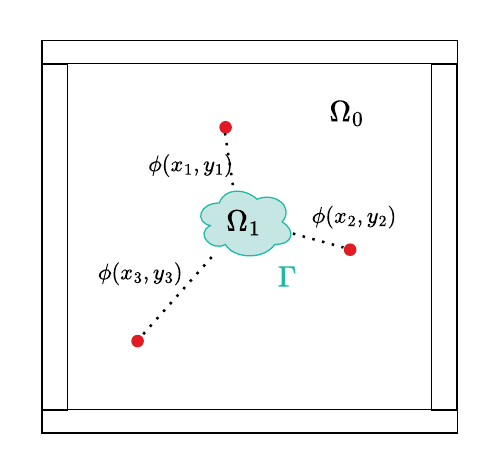}
    \caption{2D domain with an inclusion. $\phi$ is evaluated in specific points around to distinguish different geometries. It serves as input of the branch net.}
    \label{Fig:geometry_with_sdf_points}
\end{figure}

To encode the fracture geometry, the signed distance function $\phi$ is sampled at multiple points ${(\xi_i,\eta_i)}_{i=1}^N$ throughout the domain, as illustrated in Figure \ref{Fig:geometry_with_sdf_points}. The vector of values ${\phi(\xi_i,\eta_i)}$ constitutes the input of the DeepONet branch network, where $\phi(x,y)$ denotes the radial distance from $(x,y)$ on the domain to the boundary of the inclusion (negative inside, positive outside). 

Since each geometry must be represented by a unique encoding, the mapping must be injective. This requires selecting a sufficient number of sampling points to encode and discriminate among the geometries of interest. Future work may investigate how both the number and the spatial distribution of these points can be optimized in order to cover a broader class of shapes.

The trunk network receives local information at the query point $(x,y)$. Its input vector includes $(x,y)$, the local SDF value $\phi(x,y)$, and the components of its gradient $\bigl(\phi_x(x,y),\phi_y(x,y)\bigr)$. In particular, the gradient $\nabla\phi(x,y)$ yields the normal direction to the boundary when $(x,y)\in\Gamma$, which is used in enforcing the continuity of flux (zero traction) at the inclusion. Thus, the network has access to both global geometric features (via the SDF samples in the branch input) and local geometric features (via $\phi(x,y)$ and $\nabla\phi(x,y)$ in the trunk input).

Both branch and trunk inputs are processed by separate neural networks whose outputs are combined (element-wise or by inner product) to produce the final scalar output. Empirically, this separation improves generalization. In the present setting, the DeepONet output, which is the scalar field $W(x,y)$, is constrained to satisfy the Helmholtz equation. Thanks to automatic differentiation, we compute the spatial derivatives needed to evaluate the PDE residual at arbitrary points. During training, the loss function penalizes these Helmholtz residuals (and any boundary/interface condition violations) to guide the learning process with backpropagation.

Instead of introducing a volumetric source term in the PDE (which can cause large gradients and training instability), an incident wave is analytically incorporated and the network is trained to predict the scattered field $W_s$ only. This is achieved by imposing the incident wave as an external condition (effectively, a Sommerfeld radiation condition at infinity). By focusing on the homogeneous Helmholtz equation for $W_s$, we avoid the difficulties associated with a localized source in the PINN loss function.

\section{Results} \label{sec:result}

This section presents the hyperparameters setup, the training results, and a quantitative validation against FEM reference solutions.

\subsection{Setup and assumptions}

The domain is the 2D square $[0,1]\times[0,1]$, divided into two homogeneous regions: the exterior $\Omega_0$, which is filled with an elastic material (characterized by the Lamé coefficients $\lambda_0,\mu_0$ and a density $\rho_0$), and the interior inclusion $\Omega_1$, which is a void (so that $\lambda_1=\mu_1=0$ and $\rho_1=0$). The incident wave $W_{\mathrm{inc}}$ is a time-harmonic SH plane wave of frequency $\omega$. Its displacement field is 
\begin{equation}
u_{\mathrm{inc}}(\mathbf{x},t) = (0,0,w_{\mathrm{inc}}(\mathbf{x},t)), \quad w_{\mathrm{inc}}(\mathbf{x},t) =\Re\!\left(|W_{inc}|e^{i\left(k_s\,\mathbf{d}\cdot\mathbf{x}-\omega t\right)}\right), 
\end{equation}
where $|W_{inc}|$ is the incident wave amplitude, $\mathbf{d}$ the propagation direction, and $k_s$ the shear wavenumber in $\Omega_0$. We assume the inclusion boundary is perfectly rigid, so the incident wave is fully reflected. The outer boundary of the square is treated as absorbing for the scattered field $W_s$.

\begin{figure}[htbp]
    \centering
    \includegraphics[width=0.35\textwidth]{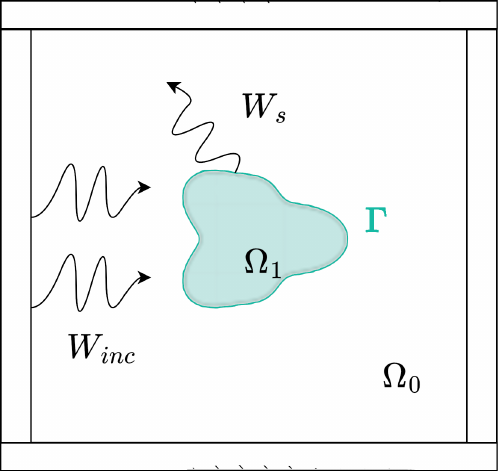}
    \caption{2D Domain used for experiment. Highlight the incident wave field $W_{inc}$ and the scattered field $W_{s}$}
    \label{Fig:domain_representation}
\end{figure}

Figure~\ref{Fig:domain_representation} shows the two-dimensional computational domain used in our experiments. The inclusion is chosen to have a concave shape to promote both constructive and destructive wave interference, causing the scattered field to remain relatively localized. This shape was also chosen to be symmetric in order to reduce the amount of distinct geometric information that must be learned, since we treat this as a proof of concept. For the same reason, the different geometries considered during training correspond to the same base geometry, simply rotated by a given angle.

For the sake of simplicity, and to facilitate both the handling of the geometry and to achieve symmetry, the inclusion boundary (plotted in Figure \ref{Fig:domain_representation}) is described in parametric form. It is given by the polar equation:

\begin{equation}
\begin{aligned}
R(\theta) &= 0.2\Big[1 + 0.08\cos(2\theta) + 0.23\cos(3\theta) + 0.11\cos(5\theta)\Big],\\
x(\theta) &= 0.5 + R(\theta)\cos(\theta),\\
y(\theta) &= 0.5 + R(\theta)\sin(\theta).
\end{aligned}    \label{eq:polarEquation}
\end{equation}
for $\theta \in [0, 2\pi)$. 

$\phi$, the radial distance, can be retrieved from the parametrization equation and can be written: 
\begin{equation} 
\phi(x,y) = r - R(\theta), \quad r = \sqrt{(x-x_c)^2 + (y-y_c)^2}, \quad \theta = \mathrm{atan2}(y-y_c,\,x-x_c), 
\end{equation}
where $(x_c,y_c)=(0.5,0.5)$ is the center of the inclusion and $R(\theta)$ is given by the polar formula given in Equation \ref{eq:polarEquation}.

This parametrization is also convenient because it prevents the spatial gradients from varying too abruptly in certain regions. The geometry features a concave region that interacts with the wavefront in a way that facilitates both constructive and destructive interference. As a result, the scattered field remains relatively localized rather than becoming overly diffuse. However, the effect of this choice on convergence has not been quantitatively assessed. At least from a qualitative standpoint, it makes the different fields easier to distinguish in the visualizations.

\begin{figure}[htbp]
    \centering
    \includegraphics[width=0.55\textwidth]{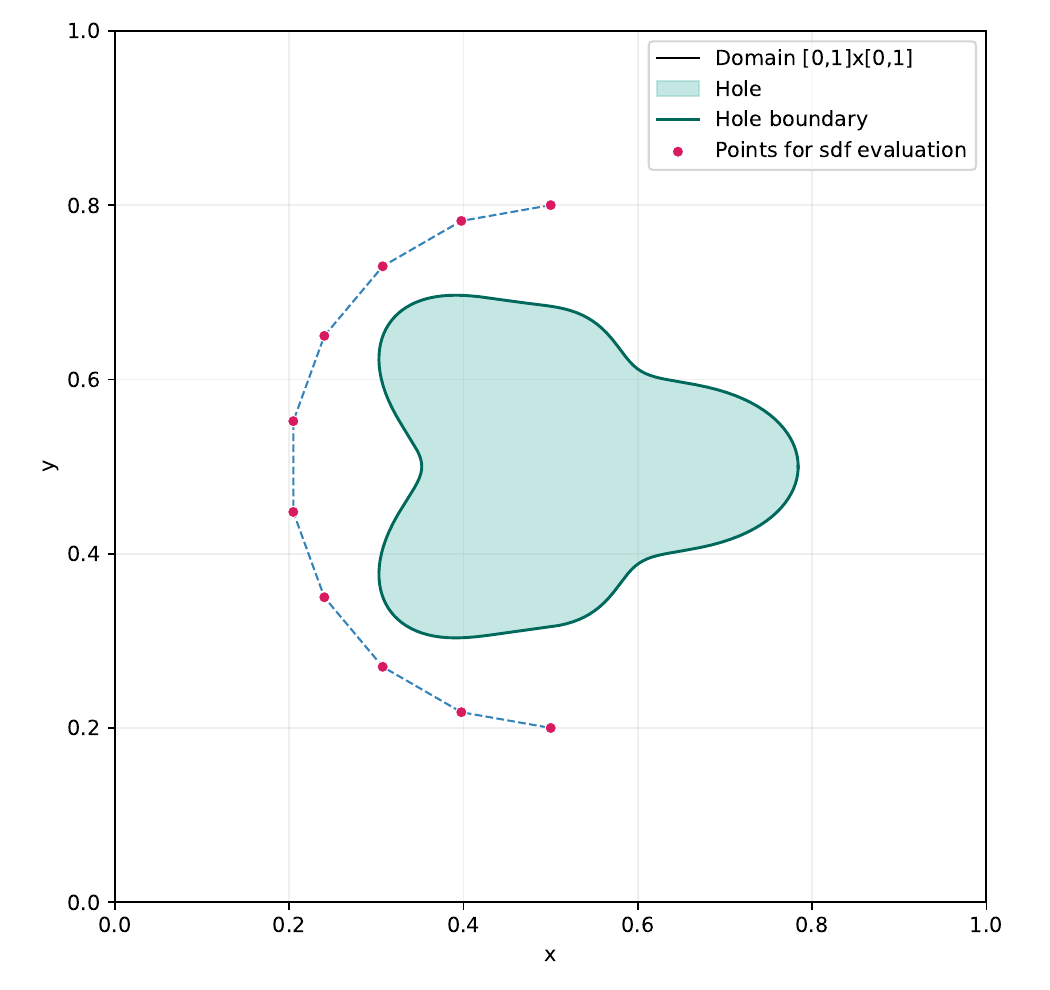}
    \caption{Symmetrical geometry encoded with $\phi$  evaluated in ten different points. It serves as input of the branch net.}
    \label{Fig:geometry_with_sdf_points_expe}
\end{figure}

As described earlier, a sufficient number of sampling points is required to discriminate the geometries of interest. In the experiments, five rotated variants of the base geometry are considered, and $N=10$ sampling points are used as shown in \ref{Fig:geometry_with_sdf_points_expe}, which is sufficient to distinguish among them. Collocation points are distributed throughout the domain; these are the points at which the network is evaluated during training. They are sampled uniformly, with a total of 15{,}000 points, and then filtered to retain only those lying in $\Omega_0$, i.e. those for which $\phi > 0$. In addition, within a thin band around the boundary defined by $\phi$, namely $\Delta\phi = 0.01$, an extra 50\% of collocation points is added. At these points, the residual of the Helmholtz equation is penalized. 20{,}000 additional points are introduced on the outer boundaries of the domain, while 2{,}000 points per geometry are distributed along the corresponding boundaries characterized by $\phi = 0$. The numbers of collocation points and boundary points were chosen empirically so as to maintain a numerical balance among the different loss terms being enforced. More optimal configurations are likely possible. The training set consists of five geometries, obtained by rotating the base geometry by five distinct angles: $-30^\circ$, $-10^\circ$, $0^\circ$, $10^\circ$, and $30^\circ$.


This work relies on the DeepONet framework implemented by DeepXDE \cite{lu2021deepxde}, using TensorFlow as the computational backend. The optimizer used in this work is the standard Adam algorithm. As described in the methodology section, the loss function is the mean squared error (MSE) designed to penalize the residual of the Helmholtz equation together with the associated boundary conditions. Both the branch net and the trunk net are standard fully connected neural networks, since architectural possibilities exploration is not the focus of this study. For both part, architectures with four layers and roughly one hundred neurons per layer were found empirically to yield convergence. This choice remains empirical, and further optimization is likely possible.


For the equation parameters, $\mu_0 = 1$ and $\rho_0 = 1$ are chosen, with no intention of matching any specific existing material. The frequency of the harmonic source, and thus the angular frequency $\omega$, is selected to remain relatively low. The goal is to keep the different terms in the equation balanced, since $\omega^2$ can very quickly dominate the residual. In real applications, however, shorter wavelengths are of greater interest, as defects are typically at least of the same order of magnitude. Investigating whether training remains convergent as the frequency increases therefore appears to be one of the most promising directions for future work.

\begin{figure}[ht]
    \centering
    \includegraphics[width=0.8\textwidth]{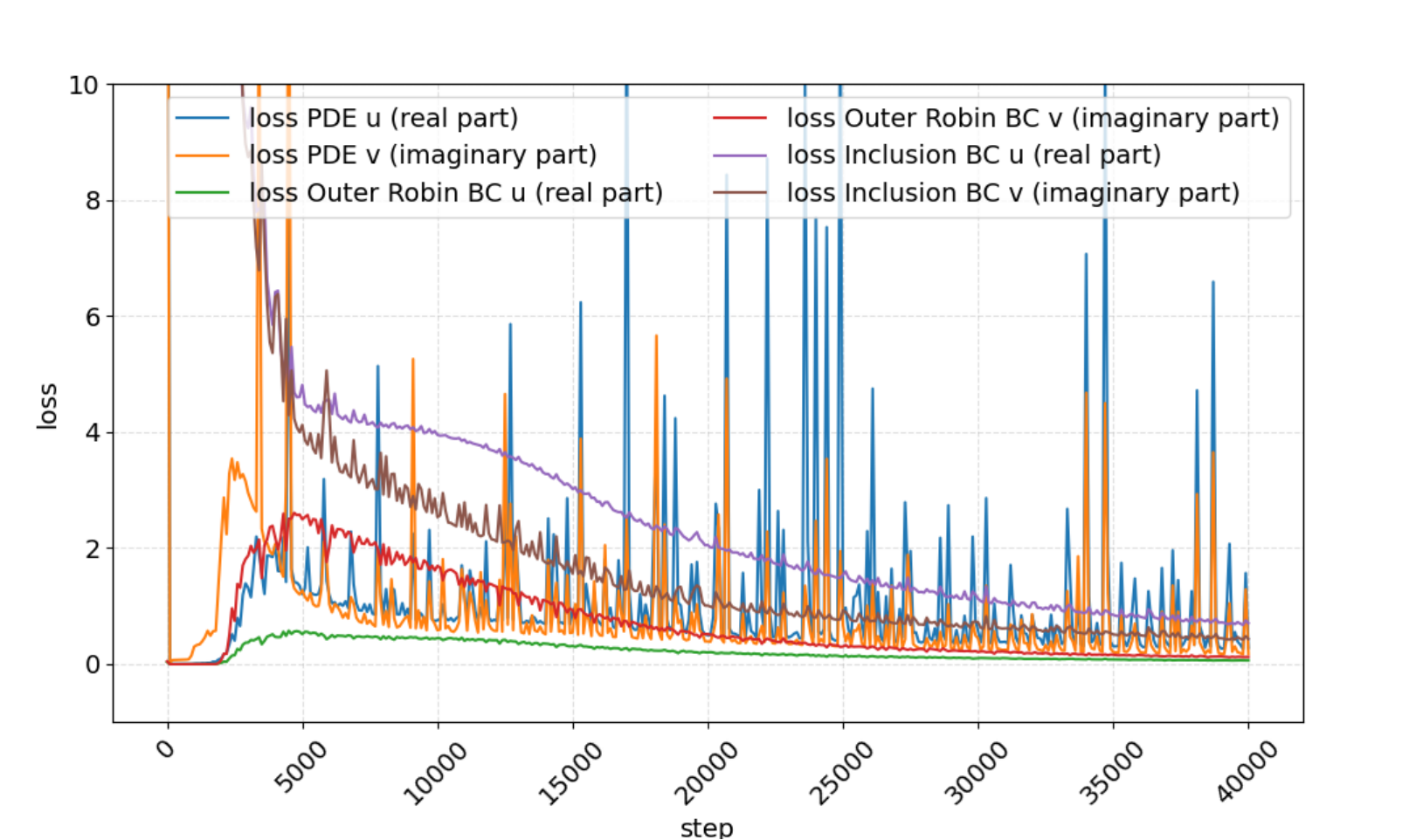}
    \caption{Evolution of the training losses over iterations. Individual contributions from the PDE residual, the inner boundary condition ($\Gamma$), and the outer absorbing boundary condition ($\Gamma_{\mathrm{out}}$) are shown separately.}
    \label{Fig:train_losses}
\end{figure}

Figure~\ref{Fig:train_losses} shows the evolution of the training losses over iterations. The total loss, as well as its individual components associated with the PDE residual, the inner boundary condition, and the outer absorbing boundary condition, all decrease steadily, indicating convergence of the training procedure.

\subsection{Finite element simulations for validation}

To assess both the accuracy and the generalization ability of the trained model, it is evaluated for geometries corresponding to every integer angle between $-60^\circ$ and $+60^\circ$. For each of these angles, the predicted values over the domain are compared with the corresponding FEM simulation results, which serve as the reference solution. The points used to compute errors are the mesh nodes from the FEM simulation; the trained model is thus evaluated at these same locations.

Validation is carried out by comparing the network predictions against FEM reference solutions computed for identical configurations: same geometry, absorbing boundary conditions on the outer boundary, material parameters, and incident wave amplitude and direction. The FEM simulations were carried out using the open-source \textsc{FreeFem++} library \cite{MR3043640}. The weak formulation of the Helmholtz problem solved by \textsc{FreeFem++} reads:

\begin{equation}
\begin{aligned}
&\int_{\Omega_0}\Big(\rho_{0}\,\omega^{2}\,W_{s}\,v-\mu_{0}\,\nabla W_{s}\cdot\nabla v\Big)\,\mathrm{d}\Omega \\
&\quad+\int_{\Gamma_{\mathrm{out}}} i\,\mu_{0}\,k_{0}\,W_{s}\,v\,\mathrm{d}\Gamma \\
&\quad-\int_{\Gamma} \mu_{0}\,i\,k_{0}\,(\mathbf{d}\cdot\mathbf{n})\,W_{\mathrm{inc}}\,v\,\mathrm{d}\Gamma
=0.
\end{aligned}
\label{eq:weak_form_ws}
\end{equation}

To ease the parametrization of the domain meshing, \textsc{Gmsh} has been used \cite{geuzaineGmsh3DFinite2009}. For this simulated example, the mesh consists of $\approx 2820$ nodes, with slight variations depending on the geometry. It is refined along the inner inclusion boundary $\Gamma$.

\begin{figure}[ht]
\centering
\begin{subfigure}{0.32\textwidth}
  \centering
  \includegraphics[width=\linewidth, trim=0 0 0 21, clip]{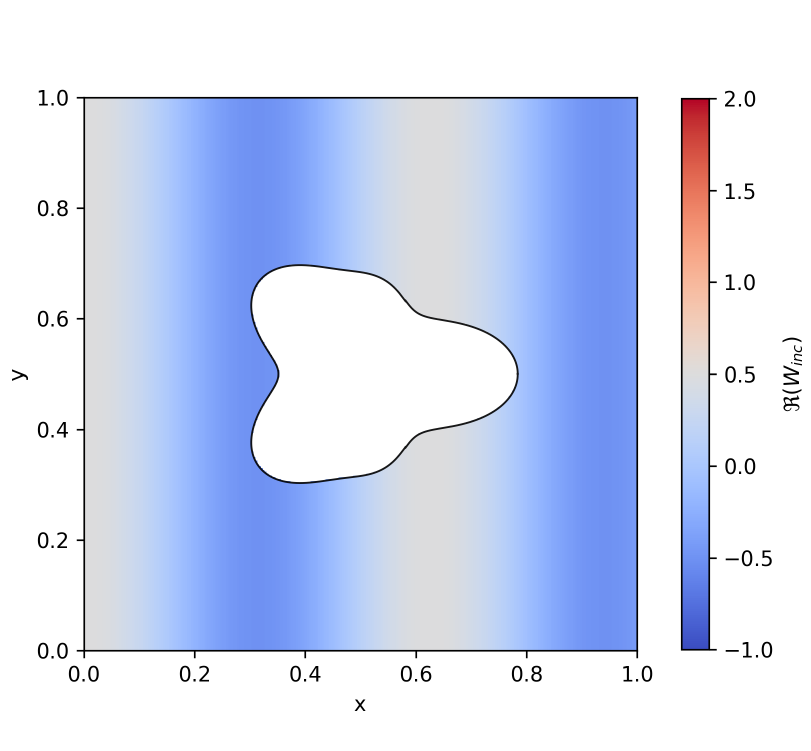}
  \caption{Incident wave}
  \label{fig:winc}
\end{subfigure}\hfill
\begin{subfigure}{0.32\textwidth}
  \centering
  \includegraphics[width=\linewidth, trim=0 0 0 21, clip]{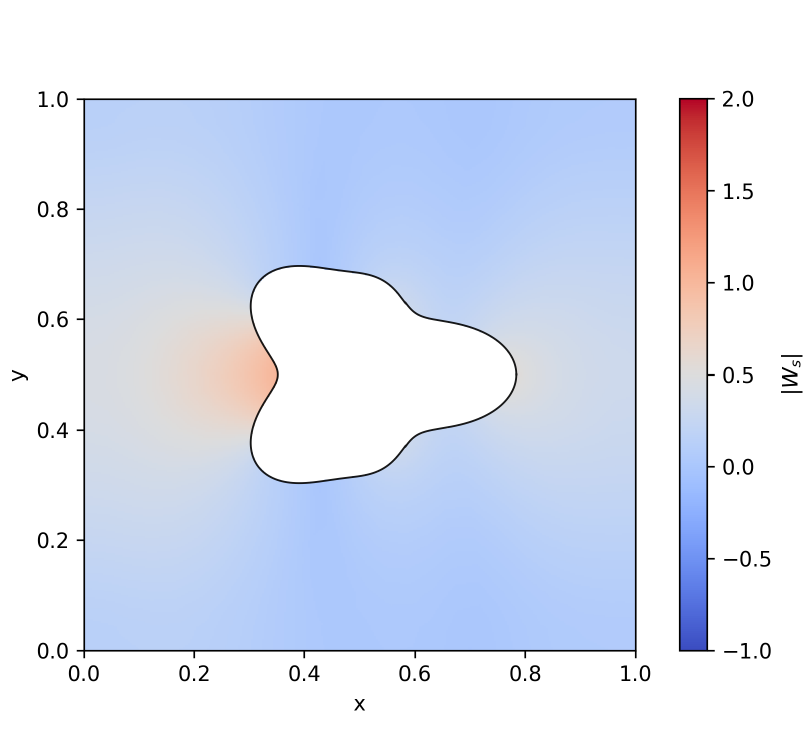}
  \caption{Scattered field amplitude}
  \label{fig:ws}
\end{subfigure}\hfill
\begin{subfigure}{0.32\textwidth}
  \centering
  \includegraphics[width=\linewidth, trim=0 0 0 21, clip]{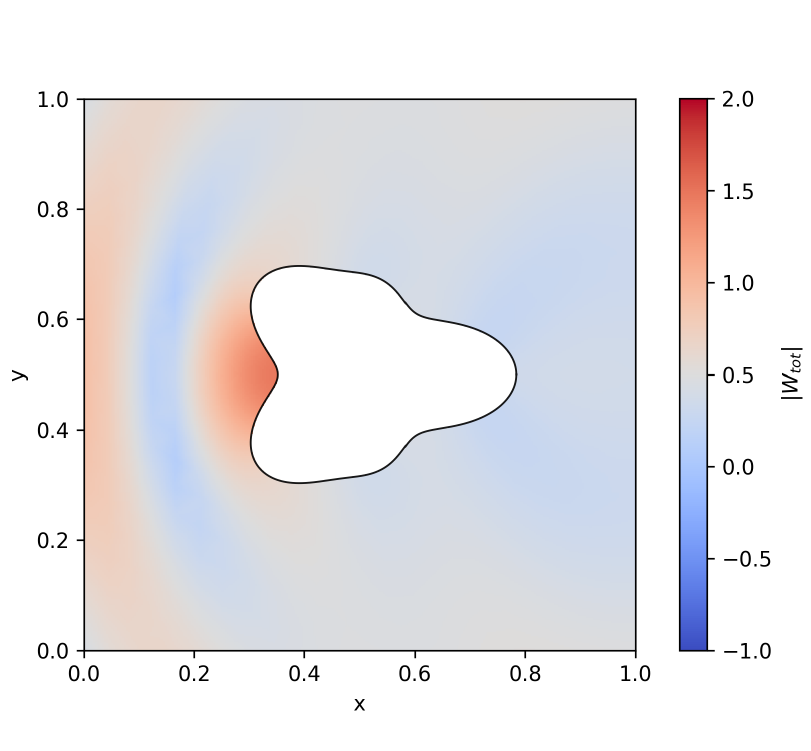}
  \caption{Total field amplitude}
  \label{fig:wtot}
\end{subfigure}
\caption{FEM solution of the Helmholtz problem (Incidence angle: $0^\circ$)}
\label{fig:fem_three_side_by_side}
\end{figure}

Figure~\ref{fig:fem_three_side_by_side} shows, side by side, the incident wave, the scattered field amplitude, and the total field amplitude for an incidence angle of $0^\circ$. Physically, $|W|$ represents the local oscillation amplitude, and therefore indicates where interference effects amplify or cancel the signal. The localized character of the scattered field, concentrated near the concavity, is consistent with the qualitative behavior expected for this type of geometry.

\begin{figure}[ht]
\centering

\begin{subfigure}{0.32\textwidth}
  \centering
  \includegraphics[width=\linewidth, trim=0 0 0 21, clip]{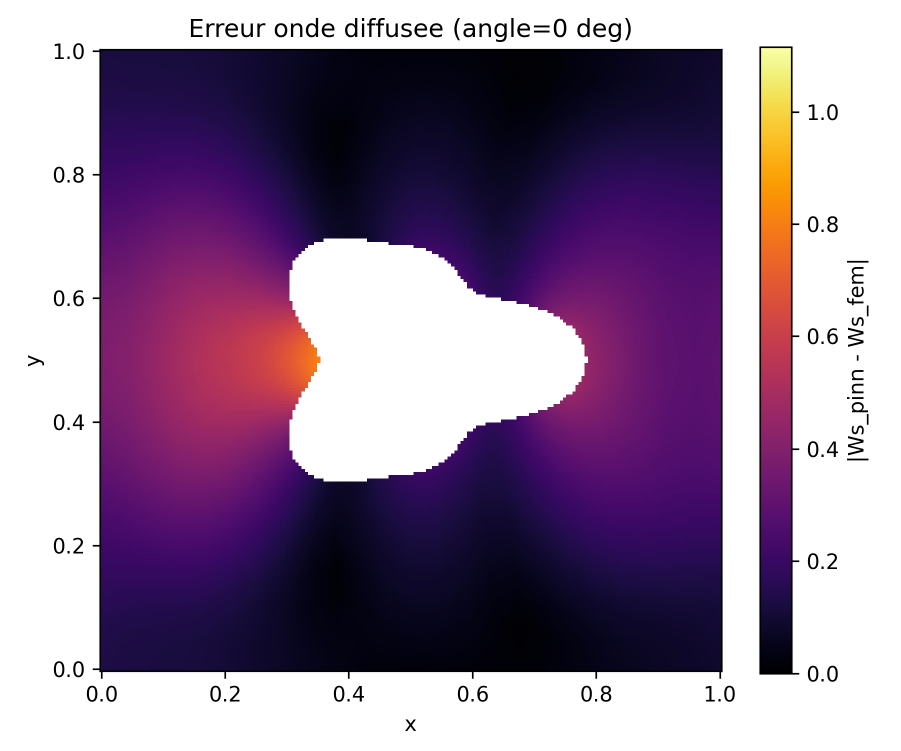}
  \caption{Rotation angle $0^\circ$}
  \label{fig:img1}
\end{subfigure}\hfill
\begin{subfigure}{0.32\textwidth}
  \centering
  \includegraphics[width=\linewidth, trim=0 0 0 21, clip]{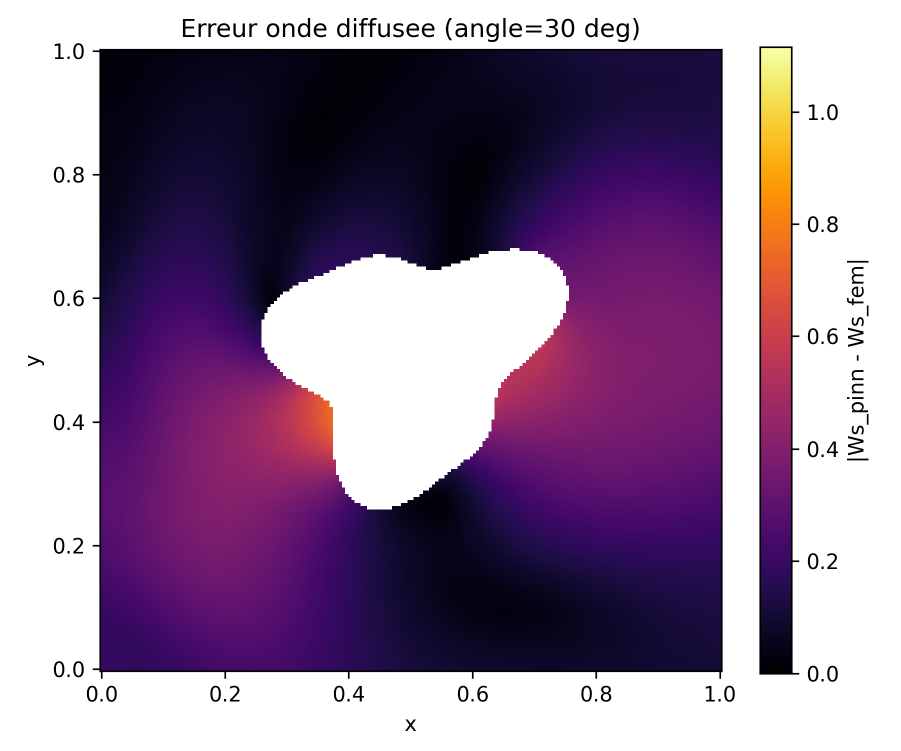}
  \caption{Rotation angle $+30^\circ$}
  \label{fig:img2}
\end{subfigure}\hfill
\begin{subfigure}{0.32\textwidth}
  \centering
  \includegraphics[width=\linewidth, trim=0 0 0 21, clip]{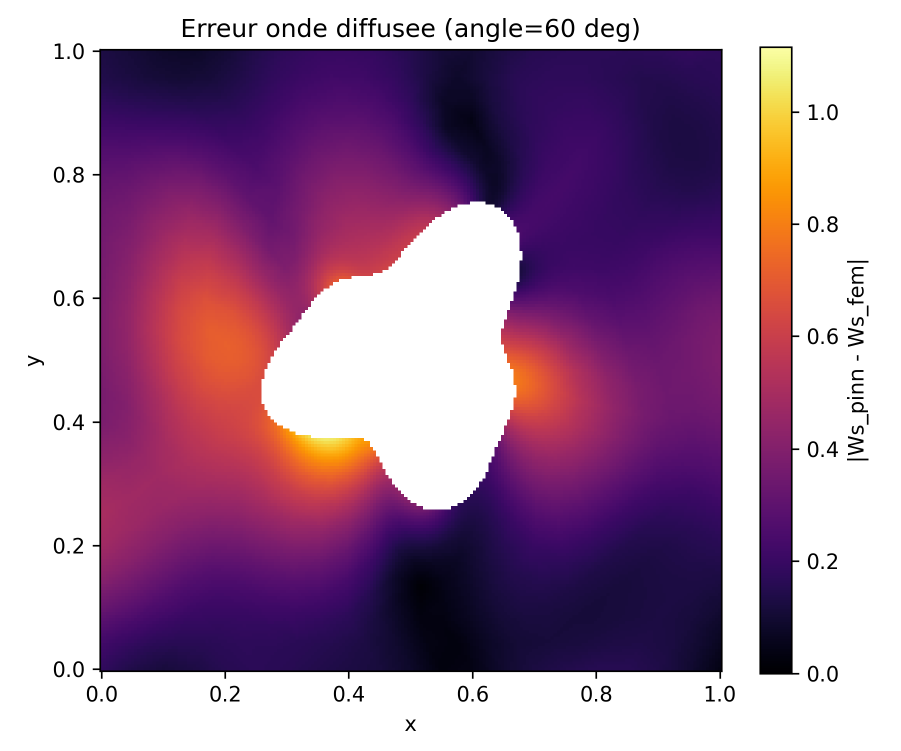}
  \caption{Rotation angle $+60^\circ$}
  \label{fig:img3}
\end{subfigure}

\caption{Absolute pointwise error between the trained model predictions and the FEM reference solutions, for three incidence angles. The $+60^\circ$ case lies outside the training range, i.e., $[-30^\circ, +30^\circ]$.}
\label{fig:three_side_by_side}
\end{figure}

Figure~\ref{fig:three_side_by_side} qualitatively illustrates the absolute pointwise error between the trained model predictions and the FEM reference solutions for the angles $0^\circ$, $+30^\circ$, and $+60^\circ$. From a qualitative standpoint, the $+60^\circ$ configuration, which lies farthest from the training geometries, exhibits the largest error.

\begin{figure}[h!]
\centering

\begin{subfigure}{0.48\textwidth}
  \centering
  \includegraphics[width=\linewidth]{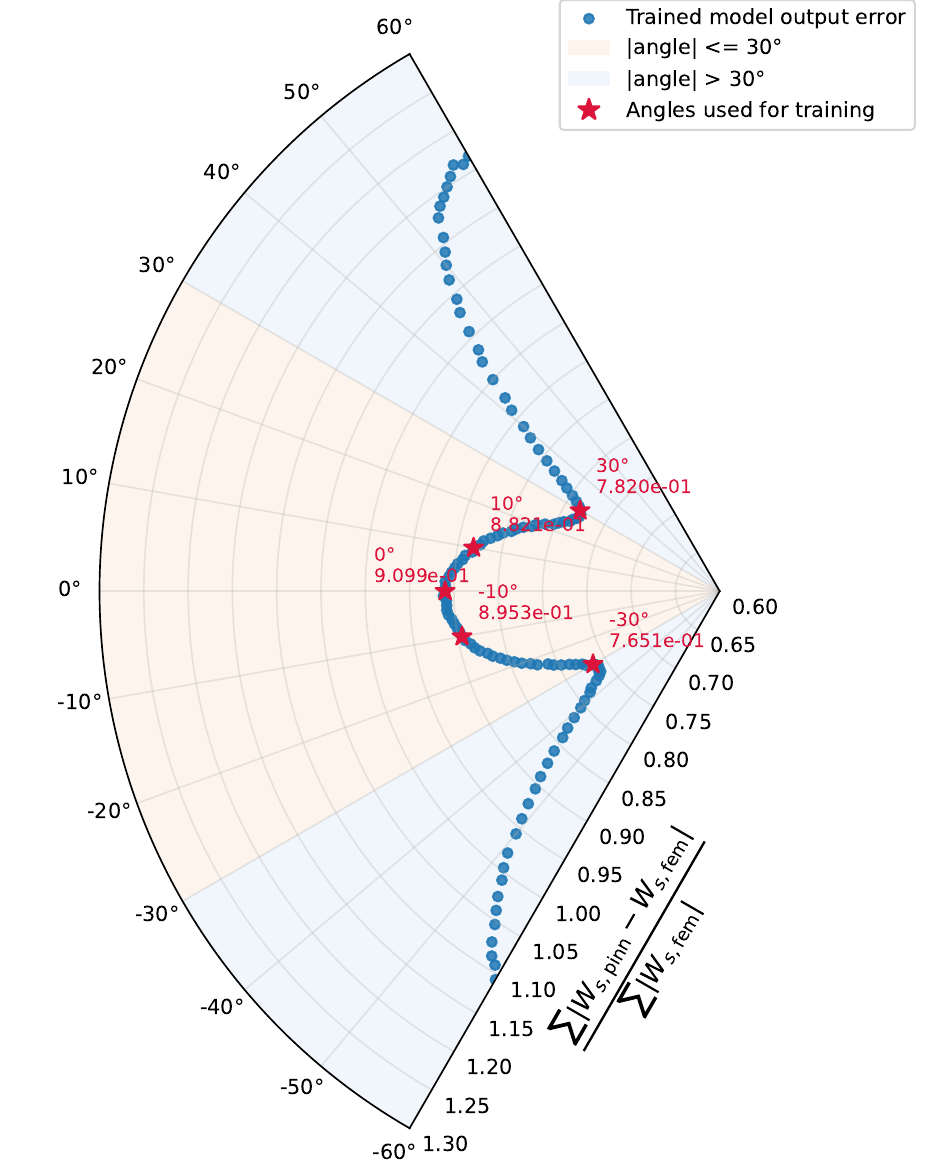}
  \caption{Error on the whole domain $\Omega_0$}
  \label{fig:error}
\end{subfigure}\hfill
\begin{subfigure}{0.48\textwidth}
  \centering
  \includegraphics[width=\linewidth]{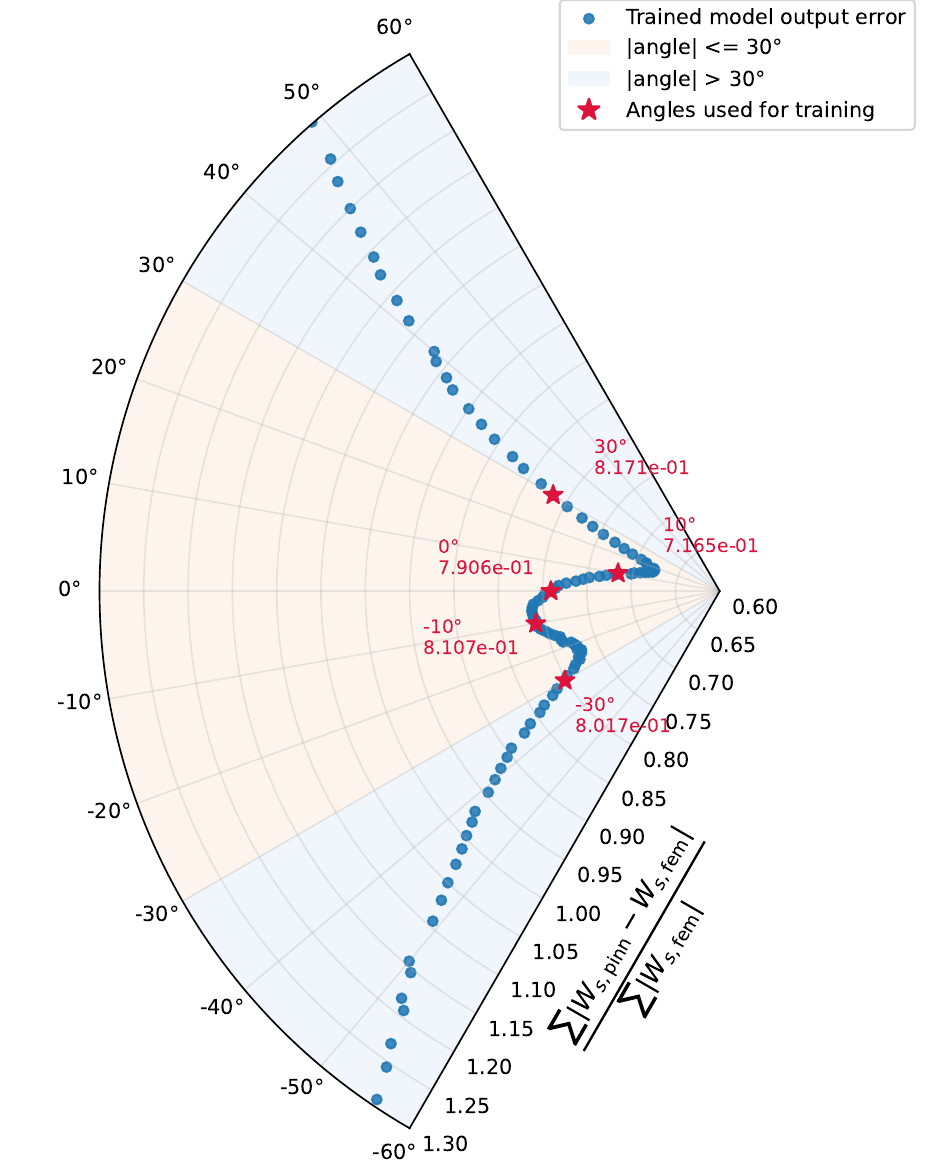}
  \caption{Error on the subdomain $\mathcal{D}_2$}
  \label{fig:error_subdomain}
\end{subfigure}

\vspace{6pt}

\begin{subfigure}{0.45\textwidth}
  \centering
  \includegraphics[width=\linewidth]{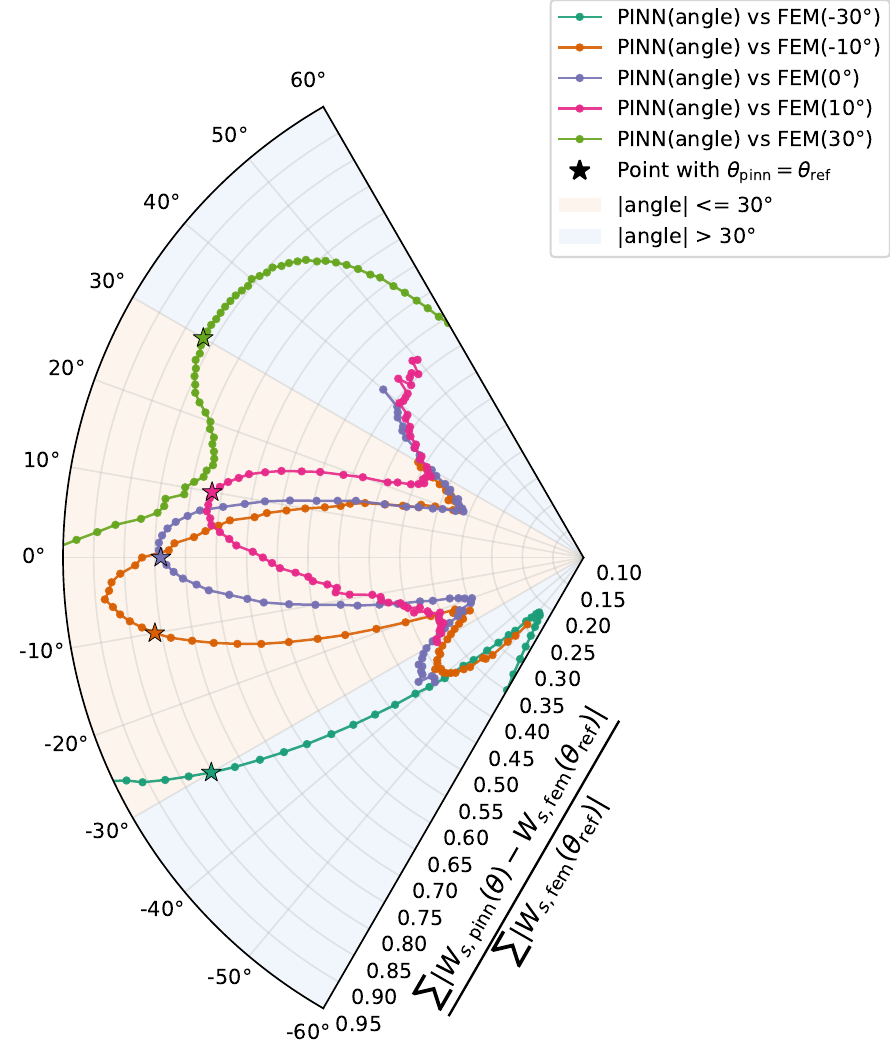}
  \caption{Normalized error with respect to each training geometry}
  \label{fig:refangle}
\end{subfigure}

\caption{Normalized pointwise error between the trained model predictions and the FEM reference solutions, as a function of the incidence angle. The vertical dashed lines at $\pm 30^\circ$ delimit the training range.}
\label{Fig:polar_plot_a}
\end{figure}

\begin{figure}[h!]
\centering
\includegraphics[width=0.45\textwidth]{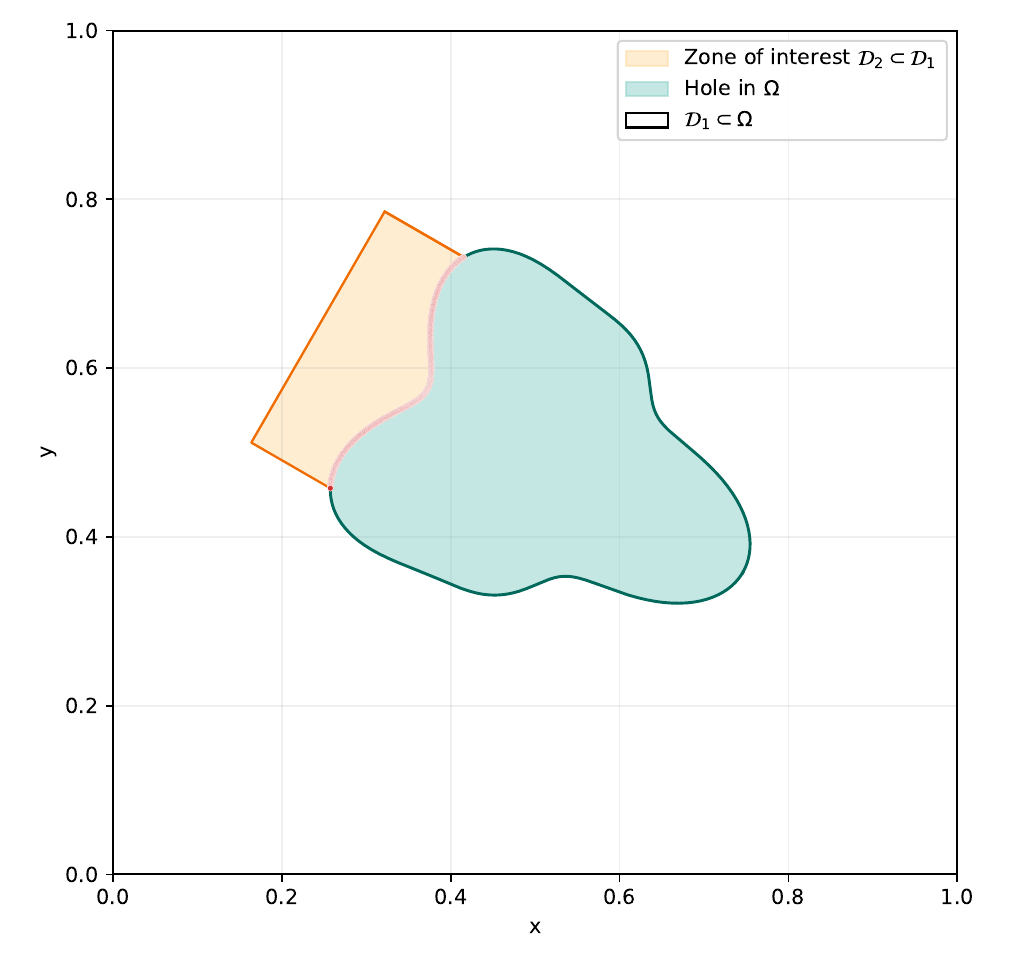}
\caption{Subdomain of interest $\mathcal{D}_2$ used to compute the localized error near the concavity.}
\label{Fig:polar_plot_b}
\end{figure}

In Figure~\ref{Fig:polar_plot_a}, subfigures~\subref{fig:error} and~\subref{fig:error_subdomain} show the normalized sum of absolute pointwise differences between the trained model and the FEM solution, reported relative to the FEM values. The first considers points over the entire domain $\Omega_0$, while the second focuses on a subregion around the concavity (shown in Figure~\ref{Fig:polar_plot_b}), where the error is most concentrated.

Once the model is applied to geometries outside the range spanned by the training set, i.e. $[-30^\circ, +30^\circ]$, the error rises sharply. By contrast, it remains controlled for angles within this interval. The error is larger on the face directly exposed to the incoming wave, particularly in the concave region (~$\mathcal{D}_2$) , where interference effects are deliberately enhanced. One possible explanation is that the network output is biased toward lower values overall, making it more difficult for the model to reproduce localized peaks of larger amplitude.

Subfigure~\ref{Fig:polar_plot_a}\subref{fig:refangle} illustrates the model’s generalization capability by comparing each evaluated geometry against every training geometry. This confirms that the model does not merely memorize training configurations, but interpolates meaningfully across the space of geometries.

\section{Conclusion and perspectives}\label{sec:persp}


This paper demonstrates the feasibility of learning the Helmholtz solution operator for non-parametric 2D geometries using a physics-informed DeepONet framework. The proposed approach aims to reconstruct the scattered field generated by a scatterer, with the Helmholtz equation serving as the sole training guide. During training, the model is presented with the same scatterer under several orientations relative to the incident wave. Its capacity to generalize is subsequently evaluated on unseen angles not included in the training set. The predictive accuracy of the learned model is finally assessed through comparison with FEM simulations.
Several natural extensions of this work are now outlined.


The main advantage of such surrogate models lies in their speed compared with FEM simulations. This makes them particularly useful in optimization loops, for instance when solving inverse problems: once the surrogate model has been trained, a large number of candidate geometries can be evaluated at a much lower computational cost, without recomputing the solution field from scratch for each configuration. This is the most direct continuation of the present work.



With the present setting, the linearity of the Helmholtz equation can be leveraged to address configurations involving multiple inclusions, following the superposition idea introduced by ~\cite{nairMultipleScatteringSimulation2025}. Unlike their approach, which is limited to a fixed configuration and does not generalize across different domains, the framework proposed in this paper is designed to be geometry-aware. A first model is trained according to the procedure described here, restricted to a single inclusion. The learned weights are then reused to initialize a second model aimed at handling multiple inclusions; in this way, the second model only has to learn the interactions between the individual scattered fields.


Another direction for future work concerns the encoding of the geometry. The proposed approach makes it possible to represent an infinite variety of geometries, unlike in \cite{nairPhysicsGeometryInformed2025}. However, since computational resources remain limited, the number and placement of points at which the SDF is evaluated should be chosen systematically, so as to capture the widest possible diversity of geometries, rather than being fixed heuristically as in the present paper.


As is common in machine learning research, empirical assessment remains essential. A wide range of questions therefore remains open regarding which layer types, network depths, and architectural choices may best promote convergence. It is also important to investigate how the training process evolves, or potentially degrades, when the frequency of the harmonic source, and thus its wavenumber, increases, and when fewer collocation and boundary points are used. It is also worth investigating whether adding further constraints could help improve convergence. Starting from the Helmholtz equation, one can derive the flux density $J(W) = \overline{W}\,\mu\nabla W - W\,\mu\nabla \overline{W}$, which satisfies the conservation law $\nabla\cdot J(W)=0$. Enforcing this relation by penalizing its residual is not equivalent to enforcing the Helmholtz equation itself. Although flux conservation carries less information than the full governing equation, it could still be beneficial as an auxiliary training constraint. In particular, it is possible to impose this conservation law without imposing the Helmholtz equation, whereas the converse is not true.



Regarding the physical modeling itself, several extensions toward more realistic configurations are worth exploring. A natural first step would be to no longer treat the inclusion as a void, but rather as a second elastic material characterized by nonzero parameters $\mu_1$ and $\rho_1$, which would introduce transmission across the interface in addition to reflection. A related question is whether it may be preferable to approximate the reflection/transmission law directly, rather than explicitly solving the field inside the inclusion. Further realism could also be achieved by incorporating volumetric and surface absorption. Finally, extending the framework to three dimensions would require carefully examining the implications for wave polarization and the associated boundary conditions.

\section*{Acknowledgement}

This work benefited from French state aid managed by the National Research Agency under France 2030 with the reference "ANR-22-PESP-0005". 

This work benefited from the GPU computing resources provided by the LAAS laboratory in Toulouse.

This work was granted access to the HPC resources of IDRIS under the allocation 2025-AD011016881 made by GENCI.

\section*{Declaration of generative AI and AI-assisted technologies in the writing process}

Generative AI tools were used to assist with translation and to improve the idiomatic quality of the English language. All scientific content was produced, reviewed, and approved by the authors, who remain fully responsible for the manuscript.

\bibliographystyle{plainnat}      
\bibliography{references}
\end{document}